\documentclass[runningheads]{llncs}
\usepackage{graphicx}

\usepackage{tikz}
\usepackage{comment}
\usepackage{amsmath,amssymb} 
\usepackage{color}
\usepackage{epsfig}
\usepackage{graphicx}
\usepackage{amsmath}
\usepackage{amssymb}
\usepackage{color,soul}
\usepackage{multirow}
\usepackage{enumitem}
\usepackage[normalem]{ulem}
\usepackage{algorithm}
\usepackage{xcolor}
\usepackage{listings}
\usepackage[font=small]{caption}
\usepackage[font=footnotesize]{subcaption}
\usepackage{stmaryrd}
\usepackage{colortbl}
\usepackage{booktabs}   
\usepackage{verbatim}
\usepackage{lipsum}
\usepackage{footnote}
\definecolor{mygray}{gray}{.9}

\newcolumntype{g}{>{\columncolor{mygray}}c}

\newcommand{\printfnsymbol}[1]{%
  \textsuperscript{\@fnsymbol{#1}}%
}
\usepackage[pagebackref=true,breaklinks=true,letterpaper=true,colorlinks,bookmarks=false]{hyperref}

\usepackage[accsupp]{axessibility}  

\begin{document}
\pagestyle{headings}
\mainmatter
\def\ECCVSubNumber{7762}  

\title{\textit{INDIGO}: Intrinsic Multimodality for Domain Generalization} 

\titlerunning{INDIGO}
%
\author{Puneet Mangla \inst{1}\thanks{Equal contribution.}\and
Shivam Chandhok \inst{3}* \and
Milan Aggarwal\inst{1} \and
Vineeth N Balasubramanian \inst{2} \and
Balaji Krishnamurthy \inst{1}
}
\authorrunning{Puneet, Shivam et al.}
%
\institute{Adobe Media and Data Science Research Lab,
Noida, India 
\and Indian Institute of Technology, Hyderabad
\and INRIA, Universite Grenoble Alpes  \\
\email{\{pmangla261, chandhokshivam, milan.ag1994\}@gmail.com, vineethnb@iith.ac.in, kbalaji@adobe.com} \\}
\maketitle

\begin{abstract}
For models to generalize under unseen domains (a.k.a domain generalization), it is crucial to learn feature representations that are domain-agnostic and capture the underlying semantics that makes up an object category. Recent advances towards weakly supervised vision-language models that learn holistic representations from cheap weakly supervised noisy text annotations have shown their ability on semantic understanding by capturing object characteristics that generalize under different domains. However, when multiple source domains are involved, the cost of curating textual annotations for every image in the dataset can blow up several times, depending on their number. This makes the process tedious and infeasible, hindering us from directly using these supervised vision-language approaches to achieve the best generalization on an unseen domain. Motivated from this, we study how multimodal information from existing pre-trained multimodal networks can be leveraged in an ``intrinsic" way to make systems generalize under unseen domains. To this end, we propose \uline{I}ntri\uline{N}sic multimodality for \uline{D}oma\uline{I}n \uline{G}eneralizati\uline{O}n (INDIGO), a simple and elegant way of leveraging the intrinsic modality present in these pre-trained multimodal networks along with the visual modality to enhance generalization to unseen domains at test-time. We experiment on several Domain Generalization settings (ClosedDG, OpenDG, and Limited sources) and show state-of-the-art generalization performance on unseen domains. Further, we provide a thorough analysis to develop a holistic understanding of INDIGO.
\keywords{Domain Generalization, Multimodality, Deep Learning}
\end{abstract}

\section{Introduction}

The underlying assumption that training and test data should comprise identically distributed samples often inhibits the applicability of deep learning models in practical scenarios where such a condition may not hold, including applications such as medical imaging, autonomous driving, robotic manipulation, etc \cite{Dou2019DomainGV,Li2021FewShotDA,DeGrave2020AIFR,Hoyer2021DAFormerIN,Wang2021DomainGF,Yue2019DomainRA}. 
Recently, the computer vision community has seen concerted efforts towards defining problem settings \cite{shu2021open,Li2021ProgressiveDE,ZeroShotDG,Mancini2020TowardsRU,Chandhok2021StructuredLE,Mangla2022COCOACA} as well as developing deep neural network models \cite{cha2021swad,rame2021ishr,EoA,Dou2019DomainGV,DGstyle,DG2021} to build systems that can learn from existing data to generalize to an unseen domain. 
Domain Generalization (DG) refers to the task of learning a model using data from source domains (for e.g. \textit{clipart}, \textit{painting}, \textit{real world}) in order to generalize and predict effectively on an unseen domain (e.g. \textit{sketch}).
Most previous approaches \cite{basicdg1,basicdg2,basicdg3,basicdg4,MTAE,DAFL,condinvariant} that address/aim to tackle the DG problem use different learning paradigms and training strategies to learn domain-agnostic semantic features that represent an object category and can thus extend to unseen domain samples at test-time. 
Other methods \cite{BalanceSpecInv,BNE,BNE2} have also shown that leveraging domain-specific features along with domain-invariant information can further improve the model's generalization on unseen domains. More recently, vision transformers (ViTs) \cite{vit,deit} have demonstrated a better ability at recognizing object shapes in less textured data such as paintings \cite{naseer2021intriguing,DGtransformer} which is a desirable trait for making models generalize to unseen domains. \par

 
An alternative strategy to address this task can be to look for other sources of information that can help disentangle domain-specific and domain-agnostic characteristics and thereby equip models with the ability to capture general domain-agnostic class-level cues. Recent progress towards weakly supervised vision-language models \cite{Radford2021LearningTV,ALBEF,li2021supervision} have shown their abilities on semantic understanding and triggered the interest in using them for practical use in various settings. These models are learned from weak supervision obtained using noisy web-based automatic label annotations and hashtags. 
However, these approaches provide a methodology for learning holistic representations from cheap, weakly supervised noisy text annotations that capture class-level semantics of object categories such as shape/content \cite{Radford2021LearningTV}. Such representations can inherently capture object characteristics that generalize to unseen domains. 
We leverage this potential of vision-language models in this work. \par  
\vspace{-5mm}
\begin{figure}
    \centering
    \includegraphics[width=0.9\textwidth]{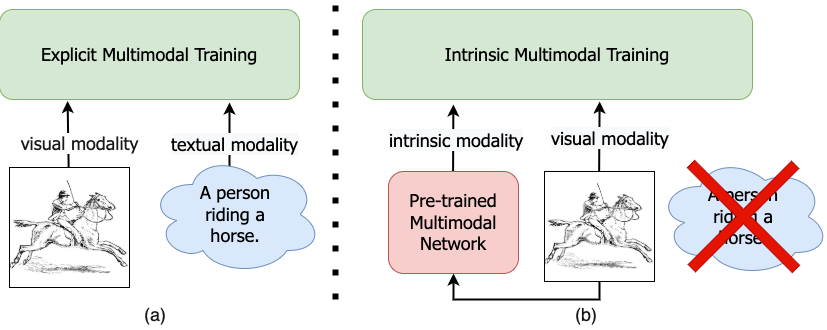}
    \caption{\scriptsize \textbf{Illustration of our broader idea.} In scenarios where we don't have access to explicit modalities like image captions for source domain data, we leverage the ``intrinsic" modality present in pre-trained multimodal networks along with visual modality obtained from image.}
    \label{fig:overview}
\end{figure}
\vspace{-10pt}
Curating semantically dense textual annotations for every image in the dataset can be a daunting task since this requires labor-intensive crowdsourcing pipelines and time. Further, when multiple source domains are involved, the annotation cost can blow up several times depending on their number, making the process tedious and infeasible. This creates a bottleneck and hinders us from directly using supervised vision-language approaches \cite{Radford2021LearningTV,li2021supervision,ALBEF} to achieve the best generalization on unseen domains. Motivated from this, we study how the multimodal information in pre-trained multimodal networks \cite{Radford2021LearningTV,li2021supervision,ALBEF} can be leveraged intrinsically to make systems robust to domain-shift and enhance generalization on unseen domains. We propose \uline{I}ntri\uline{N}sic multimodality for \uline{D}oma\uline{I}n \uline{G}eneralizati\uline{O}n (INDIGO), a simple and elegant way of leveraging the intrinsic modality present in these pre-trained multimodal networks along with the visual modality to enhance generalization to unseen domains at test-time. Figure \ref{fig:overview} provides a broader understanding of the proposed idea. To the best of our knowledge, this is the first effort to study how multimodality can be leveraged intrinsically via pre-trained multimodal models to generalize better to unseen domains. Our key contributions are as follows:
\begin{itemize}
    \item We propose \uline{I}ntri\uline{N}sic multimodality for \uline{D}oma\uline{I}n \uline{G}eneralizati\uline{O}n (INDIGO), a simple and elegant way of leveraging the intrinsic modality present in pre-trained multimodal networks along with the visual modality in order to generalize better to unseen domains. Besides that, we explore other ways of leveraging the intrinsic modality and introduce three new baselines approaches to achieve the same. We use state-of-the-art vision architectures - vision transformers (ViT) \cite{vit} - to handle the visual modality. 
   
    \item We perform comprehensive experiments on standard DG benchmarks -  DomainNet and Office-Home and show that INDIGO achieves new state-of-the-art by outperforming prior SOTAs, conventional, and newly introduced baselines. Even on more challenging settings like OpenDG and Limited Sources DG, we show INDIGO consistently outperforms aforementioned baselines.
    
    \item We perform a thorough analysis to characterize the efficacy of INDIGO in leveraging intrinsic and visual modalities obtained from pre-trained multimodal network and vision transformer (ViT), respectively.
\end{itemize}


\vspace{-7mm}
\section{Related Work}
\vspace{-5pt}
\noindent \textbf{Domain Generalization.} The reliance of deep learning models on tailored, task-specific data restricts their applicability which makes it crucial to equip these models with ability to tackle domain-shift at test-time \cite{DGSurvey1,DGSurvey2,datasetbias,bias2}. Domain Generalization (DG) \cite{basicdg1,basicdg2,basicdg3,basicdg4,MTAE,DAFL} aims to develop models that can learn from source domains (where data is abundant) and generalize to unseen novel domains given that they share same label set. Most previous approaches that tackle domain-shift learn a domain-invariant representation through data manipulation \cite{crossgrad,advaug,DGstyle,DGstyle2}, learning strategies \cite{basicdg1,basicdg2,basicdg3,basicdg4,MTAE,DAFL,condinvariant}, or optimization policies \cite{metadg1,metadg2,metadg3}. Other approaches aim to leverage domain-specific characteristics \cite{BNE,BNE2} or a balance of domain-invariant and domain-specific features to further enhance generalization on unseen domains \cite{BalanceSpecInv,Mangla2022COCOACA}. Recently \cite{shu2021open} extended conventional DG setting to an even more practical setup which allows the class label set to be disjoint for multiple source domains.
This enables practical, real-world applications by tackling cases where visual samples for all categories of interest may not be available together for all source domains due to long-tailed distributions or gradual addition of rare novel object categories.\\
\noindent \textbf{Vision Transformers.} The recent advent of attention based transformer architectures for computer vision tasks \cite{dosovitskiy2020vit,steiner2021augreg,chen2021outperform,Touvron2021TrainingDI} has motivated several efforts  to study their application for object recognition \cite{dosovitskiy2020vit,Touvron2021TrainingDI}, detection \cite{Carion2020EndtoEndOD,Zhu2021DeformableDD} and segmentation \cite{xie2021segformer,Hoyer2021DAFormerIN}. The success of these models in practical applications can be attributed to the self-attention mechanism that allows them to attend to a sequence of image patches and effectively learn global interactions better than the conv. counterparts \cite{Khan2022TransformersIV,naseer2021intriguing}. Further, these models require minimal inductive bias by design which enables them to effectively model complex functions and capture relationships from large scale datasets \cite{Khan2022TransformersIV}. These salient features allow vision transformers to perform exceptionally for computer vision tasks and better tackle nuances like occlusions, adversarial perturbations\cite{naseer2021intriguing,Naseer2021OnIA}. The capacity of transformer architectures to learn from large scale pre-training  and their ability to capture long range content dependent interactions has lead to progress towards utilising these models for processing multiple modalities for vision tasks like object detection \cite{Gupta2021TowardsGP,Kamath2021MDETRM,Maaz2021Multimodal} and classification \cite{Radford2021LearningTV,ALBEF,li2021supervision}. \\
\noindent \textbf{Multimodal Learning.} Multimodal learning aims to build models that can combine and process information from multiple
modalities like image and text. Most vision-language models use cross-modal transformers to fuse and align the information between text and image, such as LXMERT \cite{tan2019lxmert}, UNITER \cite{chen2020uniter}, ViLBERT \cite{lu2019vilbert}, VinVL \cite{zhang2021vinvl}, OSCAR \cite{li2020oscar}. Other works like ICMLM \cite{sariyildiz2020icmlm} and VirTex \cite{Desai2021VirTexLV} have shown that language supervision on COCO Captions can also produce useful visual representations. \\
However, contrastive vision-language pre-training (CLIP) \cite{Radford2021LearningTV} recently gained much attention because of its simplicity, scale, and strong results. It proposes a simple pretraining task of predicting which caption goes to which image through a image-text contrastive supervision and demonstrates that the image representations obtained are transferable to several downstream tasks like classification \cite{Radford2021LearningTV}, image retrieval \cite{Luo2021CLIP4Clip}, object detection \cite{zhou2021detecting}, image-synthesis \cite{styleCLIP}, video understanding \cite{videoclip}, 3D recognition \cite{PointCLIP}, etc. These results have garnered the attention and focus of the vision-language community to develop models using such contrastive objectives. DeCLIP \cite{li2021supervision} employs additional self-, multi-view, nearest-neighbor supervision along with image-text contrastive supervision to match the performance of CLIP but with 7.1x lesser data. ALBEF \cite{ALBEF} uses momentum distillation, a self-training method to learn from pseudo-targets produced by a  momentum model. SLIP \cite{mu2021slip} introduces a multi-task learning framework for combining self-supervised learning and CLIP pre-training. ALIGN \cite{align}, uses a larger but noisier uncurated dataset and shows similar results. \\
In this work, we leverage the intrinsic modality present in such contrastive vision-language multimodal networks as we believe their contrastive learning objective ensures that semantically similar classes representation should cluster together and different should cluster apart, allowing them to implicitly learn to focus on discriminative class-specific semantic cues of a given object category.

\section{Intrinsic Multimodality for DG}

\subsection{Background}
\noindent \textbf{Domain Generalization (DG).} The goal of DG is to learn a model using data from source domains such that it generalizes to an unseen target domain. Let $S^{Tr}=\{ (\mathbf{x}, y, d_s )| \mathbf{x} \in \mathcal{X},  y \in \mathcal{Y}, d_s \in \mathcal{D}^s\}$ denote the training set, where $\mathbf{x}$ is an image in the visual space ($\mathcal{X}$) with corresponding class label $y$ from a set of known class labels $\mathcal{Y}$ and domain label $d_s$ from a set of source domains $\mathcal{D}^s$. The test set is denoted by $S^{Ts}=\{ (\mathbf{x},  y,  d_u )| \mathbf{x} \in \mathcal{X},  y \in \mathcal{Y}, d_u \notin \mathcal{D}^s\}$ where $d_u$ represents the unseen target domain.
We aim to learn a model that captures the mapping from $\mathcal{X}\rightarrow \mathcal{Y}$ such that it is trained using $S^{Tr}$, but can predict class label $y \in \mathcal{Y}$ for an $\mathbf{x}\in \mathcal{X}$ sampled from an unseen domain $d_u \notin \mathcal{D}^s$ in $S^{Ts}$.

\noindent \textbf{Vision Transformers (ViTs).} A ViT \cite{vit} is composed of a sequence of blocks where each block contains multi-headed self-attention (\texttt{MSA}) with a feedforward network (\texttt{FFN}) and layer normalization (\texttt{LN}). An input image, $\mathbf{x}$, is first converted into a sequence of patch tokens, $\mathbf{x}_{patch}$, by dividing it with a specific patch size followed by a linear projection. Next, an additional classification (\texttt{CLS}) token, $\mathbf{x}_{\texttt{CLS}}$, is added to the sequence, followed by adding positional embedding $\mathbf{x}_{pos}$ to each token to provide positional information. All tokens are then passed through stacked transformer blocks. The \texttt{CLS} token interacts with all patch tokens and summarizes them in a single embedding vector for final classification. The processing for $k^{th}$ transformer block can be summarized as:
\begin{equation}
\begin{split}
    \mathbf{x}_0 = [\mathbf{x}_{\texttt{CLS}} \Vert \mathbf{x}_{patch}] + \mathbf{x}_{pos} \\
    \mathbf{o}_k = \mathbf{x}_{k-1} + \texttt{MSA}(\texttt{LN}(\mathbf{x}_{k-1})) \\
    \mathbf{x}_k = \mathbf{o}_k + \texttt{FFN}(\texttt{LN}(\mathbf{o}_k))
\end{split}
\end{equation}



\subsection{Motivation: Multimodal networks generalize better to different domains }

Most methods that tackle the domain-shift problem devise learning strategies that capture domain-agnostic features. Such methods work based on the assumption that a domain-invariant manifold exists where the object images lie irrespective of the domain in which they are represented.
For e.g class-level semantic cues such as  \texttt{long neck}, \texttt{long legs}, \texttt{has spots} are stable characteristic features that define the class giraffe. Hence, it is crucial to design models that focus on underlying semantic features that make up an object category and are robust to domain variations.\par

\begin{figure}
    \centering
    \includegraphics[width=1.\textwidth]{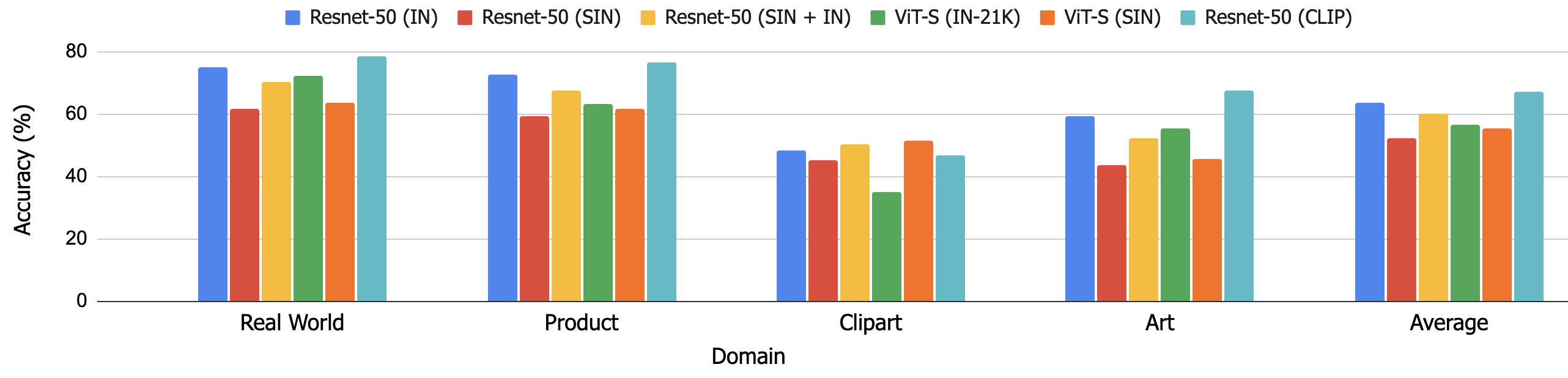}
    \caption{\footnotesize{\textbf{Generalization to different domains.} Performance of various pre-trained models on different domains of Office-Home dataset. Results are averaged over 5 runs. Vision-language model, Resnet-50 (CLIP) can been seen to outperform others at generalizing to different domains. \textbf{IN}: ImageNet-1K, \textbf{IN-21K}: ImageNet-21K, \textbf{SIN}: Stylized ImageNet.}}
    \vspace{-8mm}
    \label{fig:generalization_graph}
\end{figure}

Text can describe images with  syntactically and semantically meaningful  sentences, offering a better way to summarize their content than one-hot or soft-label vectors. 
Vision-language models  like CLIP \cite{Radford2021LearningTV} which are trained on noisy weakly aligned image-text pairs with minimal supervision are better at understanding image content in different domains as compared to other vision models (Resnet-50 and ViT-S \cite{naseer2021intriguing}) trained on ImageNet-1K/-21K \cite{deng2009imagenet}, and Stylized-ImageNet \cite{stylized-imagenet}(Figure  \ref{fig:generalization_graph}). Since these models are trained with a contrastive learning objective that implicitly encodes information about inter-class relationships, they develop the ability to focus on class-specific semantic cues (rather than texture) that help them generalize to domain shifts. \\
However, annotating every image with captions in source domains can be a daunting task because of time and labor. Motivated from this, we propose \uline{I}ntri\uline{N}sic multimodality for \uline{D}oma\uline{I}n \uline{G}eneralizati\uline{O}n (INDIGO) that exploit the large-scale pre-trained vision-language models \cite{Desai2021VirTexLV,li2021supervision,ALBEF}, by integrating the intrinsic modality present in their representations with the visual modality obtained from a vision transformer trained on source domains.
\vspace{-3mm}
\subsection{INDIGO: Leveraging intrinsic modality present in MViTs } 
\begin{figure}
    \centering
    \includegraphics[width=\textwidth]{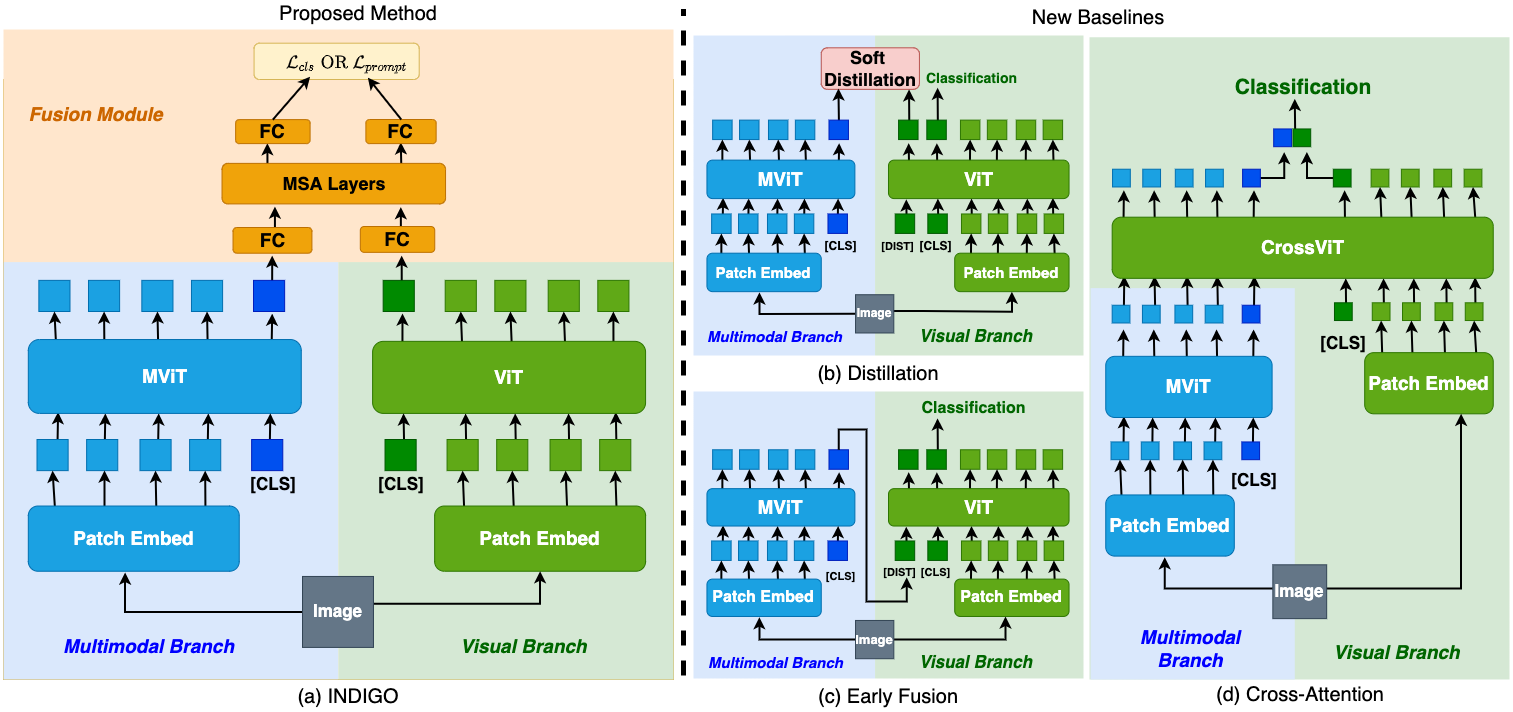}
    \caption{\scriptsize (a) (Proposed approach) \textbf{INDIGO} consists of a multimodal branch comprised of pre-trained MViT to obtain intrinsic modality , a visual branch to extract visual modality and a fusion module to combine both; (New Baselines) (b) \textbf{Distillation} considers MViT as teacher and distills a ViT with a soft distillation loss via \texttt{DIST} token; (c) \textbf{Early Fusion} fuses the intrinsic modality via \texttt{DIST} token in the input layer of the visual branch itself; (d) \textbf{Cross-Attention} uses a CrossViT \cite{chen2021crossvit} to cross-attend MViT features with ViT features.}
    \label{fig:indigo}
    \vspace{-6mm}
\end{figure}

\vspace{-3mm}
As depicted in Figure \ref{fig:indigo}a, there are three main components in our approach: (1) a \textit{multimodal branch} which consists of a multimodal vision transformer (MViT) pre-trained on image-text pairs used to extract the intrinsic modality present in it; (2) a \textit{visual branch}, which trains a vision transformer (ViT) to extract visual modality that will encode meaningful shape-biased concepts from the source domains, useful for generalization; and (3) a \textit{fusion module} which combines best of both - intrinsic and visual modality through a multi-headed self-attention mechanism for final classification. \\
\noindent \textbf{Multimodal branch.} We leverage pre-trained large-scale vision-language networks like CLIP \cite{Radford2021LearningTV}, DeCLIP \cite{li2021supervision}, ALBEF \cite{ALBEF} that use a contrastive objective to push the embeddings of matched image-text pairs together and non-matched pairs apart. The pipeline generally consists of an image encoder $f^{M}(.)$ (in our case a ViT which we call MViT), a text encoder $g(.)$, and linear projection layers $h^I(.)$ and $h^T(.)$. The image and text features (obtained from their respective encoders) are projected to the same dimension, normalized, and then aligned using the following contrastive loss: 
\begin{gather}
\nonumber 
    \mathbf{z}_i^I = \frac{h_I(f^{M}_\texttt{CLS}(\mathbf{x}_i))}{\Vert h_I(f^{M}_\texttt{CLS}(\mathbf{x}_i)) \Vert_2}; \  \mathbf{z}_i^T = \frac{h_T(g(\mathbf{t}_i))}{\Vert h_T(g(\mathbf{t}_i)) \Vert_2}\\
    \nonumber  \mathcal{L}_I = - \frac{1}{N} \sum_{i=1}^N \text{log} \frac{\text{exp}(\text{sim}(\mathbf{z}_i^I, \mathbf{z}_i^T)/\tau)}{\sum_{j=1}^N \text{exp}(\text{sim}(\mathbf{z}_i^I, \mathbf{z}_j^T)/\tau) } \\ \nonumber 
    \mathcal{L}_T = - \frac{1}{N} \sum_{i=1}^N \text{log} \frac{\text{exp}(\text{sim}(\mathbf{z}_i^T, \mathbf{z}_i^I)/\tau)}{\sum_{j=1}^N \text{exp}(\text{sim}(\mathbf{z}_i^T, \mathbf{z}_j^I)/\tau) } \\ \nonumber 
    \mathcal{L}_{contrastive} = (\mathcal{L}_I + \mathcal{L}_T)/2
\end{gather}
\noindent here $(\mathbf{x}_i$, $\mathbf{t}_i)$ denote the $i^{th}$ image-text pair in a batch of size $N$. $f^{M}_\texttt{CLS}(.)$ represents the MViT's representation corresponding to the \texttt{CLS} token. The similarity function \texttt{sim(,)} is measured by dot product, and $\tau$ is a learnable temperature variable to scale the logits. \par

In scenarios where we do not have direct access to text annotations, we can assume that an image's unnormalized projected embedding $h^I(f^{M}_\texttt{CLS}(\mathbf{x}_i))$ would be weakly aligned with its hypothetical text description. This allows us to leverage the intrinsic modality present in a pre-trained multimodal vision transformer. Hence, we propose to use this unnormalized projected embedding $h^I(f^{M}_\texttt{CLS}(\mathbf{x}_i))$ as a ``intrinsic" modality in our overall pipeline. 


\noindent \textbf{Visual branch.} The visual branch is a sibling to the multimodal branch. We employ a trainable vision transformer, $f^V(.)$, to learn visual concepts from source domains that might be absent in MViT representations but are relevant to the task. These concepts can be dataset-, domain-, or even class-specific, which, when combined with the ``intrinsic"  modality, can help boost the overall performance on the given task. Moreover, by design, since ViTs are better than CNNs in recognizing object shapes \cite{naseer2021intriguing,DGtransformer}, we believe their shape-biased representations $f^V_\texttt{CLS}(\mathbf{x})$  will further assist our overall pipeline in generalizing to unseen domains (as we show through our experiments).\\
\noindent \textbf{Fusion module.} The purpose of the fusion module is to fuse the ``intrinsic” modality $h^I(f^{M}_\texttt{CLS}(\mathbf{x})))$ (obtained from the multimodal branch) and the visual modality $f^V_\texttt{CLS}(\mathbf{x})$ (obtained from the visual branch) to perform the final classification.  We first project both of them to same space via linear projections $w^{M}(.)$ and $w^V(.)$ to obtain intrinsic modality $w^{M} (h^I(f^{M}_\texttt{CLS}(\mathbf{x})))$ and visual modality $w^V (f^V_\texttt{CLS}(\mathbf{x}))$ tokens. This is followed by a series of $K$ multi-headed self-attention blocks (\texttt{MSA}) and feed-forward networks  (\texttt{FFN}) to perform inter-modality attention on both tokens as follows 
\vspace{-7.1pt}
\begin{gather}
\nonumber \mathbf{x}_0^{M} = w^{M} (h^I(f^{M}_\texttt{CLS}(\mathbf{x}))); \  \mathbf{x}_0^{V} = w^V (f^V_\texttt{CLS}(\mathbf{x})) \\
    \nonumber \mathbf{x}_0 = [\mathbf{x}_0^{M} \ || \ \mathbf{x}_0^{V} ] \\
    \mathbf{o}_k = \mathbf{x}_{k-1} + \texttt{MSA}(\texttt{LN}(\mathbf{x}_{k-1})) \\ \nonumber 
    \mathbf{x}_k = \mathbf{o}_k + \texttt{FFN}(\texttt{LN}(\mathbf{o}_k)) \\ \nonumber 
    \mathbf{x}_K = [\mathbf{x}_K^{M} \ || \ \mathbf{x}_K^{V} ]
\end{gather}
The attention mechanism allows the intrinsic modality token to attend with the visual modality token and incorporate any dataset, domain, or class-specific concepts present in it. Similarly, the visual modality token will interact with the intrinsic modality token to learn multimodal concepts present in it. This ensures that final representations leverage the best of both modalities. Finally, the transformed representation of intrinsic modality ($\mathbf{x}_K^{M}$) is passed through a linear layer $c^M(.)$ to get class predictions and minimize cross-entropy loss. In addition to this, we add a regularizer that also minimizes classification loss on the transformed representation of visual modality token (by passing $\mathbf{x}_K^{V}$ through another linear layer $c^V(.)$). Overall loss can be written as follows
\vspace{-5pt}
\begin{gather} 
\nonumber
    \hat{y}_{M} = c^M(\mathbf{x}_K^{M}); \ \hat{y}_{V} = c^V(\mathbf{x}_K^{V})\\
    \mathcal{L}_{cls} = \lambda \cdot \mathcal{L}_{CE}(\hat{y}_{M}, y) + (1-\lambda) \cdot 
    \mathcal{L}_{CE}(\hat{y}_{V}, y)
    \label{eq:lce_loss}
\end{gather}
\noindent where $\lambda$ is the regularization hyperparameter. 
In scenarios like OpenDG, where each source domain holds disparate label sets, chances of learned representations becoming domain biased are high. Rather than minimizing Equation \ref{eq:lce_loss}, we minimize the following semantic alignment loss
\begin{gather}
    \nonumber \mathbf{z}^M = \frac{p^M(\mathbf{x}_K^{M})}{\Vert p^M(\mathbf{x}_K^{M}) \Vert_2}; \ \mathbf{z}^V = \frac{p^V(\mathbf{x}_K^{V})}{\Vert p^V(\mathbf{x}_K^{V}) \Vert_2} \\
    \nonumber \hat{y}_{M} = \left [ \frac{\text{exp}(\text{sim}(\mathbf{z}^M, g(\mathbf{t}_1))/\tau)}{\sum_{i=1}^C \text{exp}(\text{sim}(\mathbf{z}^M, g(\mathbf{t}_i))/\tau) }, ... \frac{\text{exp}(\text{sim}(\mathbf{z}^M, g(\mathbf{t}_C))/\tau)}{\sum_{i=1}^C \text{exp}(\text{sim}(\mathbf{z}^M, g(\mathbf{t}_i))/\tau) }\right ] \\
    \hat{y}_{V} = \left [ \frac{\text{exp}(\text{sim}(\mathbf{z}^V, g(\mathbf{t}_1))/\tau)}{\sum_{i=1}^C \text{exp}(\text{sim}(\mathbf{z}^V, g(\mathbf{t}_i))/\tau) }, ... \frac{\text{exp}(\text{sim}(\mathbf{z}^V, g(\mathbf{t}_C))/\tau)}{\sum_{i=1}^C \text{exp}(\text{sim}(\mathbf{z}^V, g(\mathbf{t}_i))/\tau) }\right ] \\
    \nonumber \mathcal{L}_{prompt} = \lambda \cdot \mathcal{L}_{CE}(\hat{y}_{M}, y) + (1-\lambda) \cdot 
    \mathcal{L}_{CE}(\hat{y}_{V}, y)
\end{gather}
\noindent where $p^M(.)$ and $p^V(.)$ are semantic projection layers, $\tau$ is a learnable temperature variable to scale the logits, $\mathbf{t}_i$ is the text prompt for $i^{th}$ class i.e \texttt{"a photo of \{class\}"}, and $g(.)$ is the text encoder of the pre-trained multimodal network. Enforcing that images align with their corresponding class prompts ensures that the representations do not get biased towards domains and capture domain-agnostic class-specific semantics described via class prompts.
\vspace{-10pt}
\subsection{Baselines: Other approaches for leveraging intrinsic modality}
\vspace{-2pt}
Besides proposing INDIGO, we also explore other ways to leverage the intrinsic modality obtained from the multimodal branch and combine it with the visual modality extracted from the visual branch. We introduce three baseline approaches (variations to our proposed approach) to achieve the same.\par

\noindent \textit{Logit Distillation.} As illustrated in Figure \ref{fig:indigo}b, we consider the pre-trained MViT as a teacher network and use a soft-distillation strategy to distill intrinsic modality present in it via an additional distillation (\texttt{DIST}) token similar to DeiT \cite{deit}.

\noindent \textit{Early Fusion.} Instead of using \texttt{DIST} token for performing logit distillation, we can use it to fuse the intrinsic modality in the input layer of the visual branch itself. This is illustrated in Figure \ref{fig:indigo}c. The \texttt{CLS} token $\mathbf{x}_\texttt{CLS}$ can now interact with both - intrinsic modality (provided via \texttt{DIST} token) and patch tokens $\mathbf{x}_{patch}$ to summarize the information present in them for final classification. 

\noindent \textit{Cross-Attention.} We can cross-attend features (corresponding to image patches and \texttt{CLS} token) extracted from vision-language model with image patch embeddingss $\mathbf{x}_{patch}$ and \texttt{CLS} token $\mathbf{x}_\texttt{CLS}$. For this purpose, we now employ a Cross-attention Vision Transformer (CrossViT) \cite{chen2021crossvit} rather than vanilla ViT \cite{vit} in the visual branch. This is illustrated in Figure \ref{fig:indigo}d.\par
\vspace{-5pt}
\section{Experiments and Analysis}
\vspace{-3pt}
\noindent \textbf{Closed Domain Generalization} We first perform experiments on the following domain generalization datasets under closed setting - (1) DomainNet \cite{domainnet}, a large scale dataset containing 586,575 examples from 345 classes and six domains (\textit{clipart}, \textit{infograph}, \textit{painting}, \textit{quickdraw}, \textit{real}, \textit{sketch}); and (2) Office-Home \cite{office-home}, containing 15,588 examples from 65 classes and four domains (\textit{art}, \textit{clipart}, \textit{product}, \textit{real}). \\
\noindent \textit{(\textbf{Baselines})} We evaluate and compare four kinds of training pipelines - (1) CNNs, which include state-of-the-arts \cite{cha2021swad,rame2021ishr,EoA} that use a Resnet-50 backbone; (2) ViTs, which include DeiT-S \cite{deit} (considered equivalent to Resnet-50) backbone trained in AGG manner; (3) MViTs, which include conventional ways (like zero-shot inference, transfer learning using linear layer and attention layers) of using the pre-trained MViT; and (4) MViTs + ViTs, that include our newly introduced fusion baselines (distillation, early fusion, and cross attention) and our proposed fusion, INDIGO (all using a DeiT-S visual backbone). Implementation and architectural details of fusion module are described in the supplementary.\\
\noindent \textit{(\textbf{Training and evaluation protocol})}  Following previous works \cite{cha2021swad,rame2021ishr,domainbed}, we consider each domain as the target domain and the rest domains as source domains for training. We use test-domain validation (reporting best performance on test set) and training-domain validation model selection criteria (using a validation set) for DomainNet and Office-Home, respectively, as described in \cite{domainbed}. \\
\noindent \textit{(\textbf{Results})}  Table \ref{table:clip_dg_results} presents our results when CLIP-ViT-B/16 \cite{Radford2021LearningTV} is used as an MViT in the multimodal branch. As we can see, INDIGO achieves new state-of-the-art results by outperforming all the compared methods by good margins. In particular, on challenging domains like  \textit{quickdraw} where conventional ways of using MViTs perform worse than prior arts, INDIGO achieves the best performance by leveraging the best of both  - intrinsic and the visual modality. Further, we can observe that ViTs trained with simple vanilla AGG loss easily beat state-of-the-art CNN-based approaches - SWAD \cite{cha2021swad}, EoA \cite{EoA}. This shows that their design offers shape-biased representations (compared to CNNs), which INDIGO leverages. Amongst our newly proposed baselines, Early Fusion stands out as the best competition. \\
\begin{table}[t] 
\caption{\scriptsize \textbf{ClosedDG results.} Performance of INDIGO on DomainNet (\textbf{C}: clipart, \textbf{S}: sketch, \textbf{P}: painting, \textbf{Q}: quickdraw, \textbf{I}: infograph) and Office-Home (\textbf{R}: real world, \textbf{C}: clipart, \textbf{P}: product, \textbf{A}: art)  datasets under closed setting. We highlight the \textbf{best results} and the \uline{second best} results. The results are averaged over five runs. INDIGO achieves new state-of-the-art by outperforming all compared methods by good margins.}
\begin{center}
\setlength{\tabcolsep}{5pt}
\resizebox{\textwidth}{!}{
\begin{tabular}{c|c|c|c|c|c|c|g|c|c|c|c|g}
\toprule[0.4mm]
\multirow{2}{*}{\textbf{Type}} & \multicolumn{1}{c|}{\multirow{2}{*}{\textbf{Method}}}                     & \multicolumn{6}{c|}{\textbf{DomainNet}}                                                                                                                                                 & \multicolumn{5}{c}{\textbf{Office-Home}}                                                                                                             \\ 
                                       & \multicolumn{1}{c|}{}                                                     & \multicolumn{1}{c|}{\textbf{C}} & \multicolumn{1}{c|}{\textbf{S}} & \multicolumn{1}{c|}{\textbf{P}} & \multicolumn{1}{c|}{\textbf{Q}} & \multicolumn{1}{c|}{\textbf{I}} & \textbf{Avg.} & \multicolumn{1}{c|}{\textbf{R}} & \multicolumn{1}{c|}{\textbf{C}} & \multicolumn{1}{c|}{\textbf{P}} & \multicolumn{1}{c|}{\textbf{A}} & \textbf{Avg.} \\ \midrule
\multirow{20}{*}{CNNs}                  & \multicolumn{1}{c|}{AGG}  & 58.4 & 49.9 & 47.3 & 13.4 & 19.8  & 37.76 & 77.3 & 53.4 & 76.5 &  62.7 &  67.47   \\
                                        & \multicolumn{1}{c|}{IRM \cite{arjovsky2019invariant}}  & 51.0 & 44.7 & 38.8 & 11.8 & 16.7 & 32.6 & 77.2 & 52.3 & 75.2 & 61.8 & 66.63        \\
                                        & \multicolumn{1}{c|}{DRO \cite{DRO}}  & 47.8 & 40.7 & 36.3 & 9.0 & 17.2 & 30.2 & 77.7 & 52.9 & 75.5 & 61.6 & 66.93        \\
                                        & \multicolumn{1}{c|}{Mixup \cite{zhang2018mixup}}  & 55.8 & 49.2 & 46.2 & 12.8 & 19.2 & 36.64  & 79.2& 54.7 & 77.3 & 64.7&    68.98   \\
                                        & \multicolumn{1}{c|}{MLDG \cite{metadg1}}  & 59.3 &  51.2 & 48.8 & 14.0 & 20.3 & 38.72 & 78.6 & 54.5 & 75.9 & 63.7 &    68.18    \\
                                        & \multicolumn{1}{c|}{CORAL \cite{dcoral}}  & 58.8 & 50.8 & 47.5 & 13.6& 20.8 & 38.3 & 77.9 & 55.3 & 76.7 & 64.4 &   68.58     \\
                                        & \multicolumn{1}{c|}{MMD \cite{DAFL}}  & 54.6 & 47.5 & 44.9 & 12.6 & 19.6 & 35.84 & 78.1 & 53.7 & 76.1 & 63.0 & 67.73        \\
                                        & \multicolumn{1}{c|}{DANN \cite{ganin2016domain}}   & 53.8 & 46.7 & 43.5 & 11.8 & 17.5 & 34.66 & 76.6 & 51.7 & 74.1 & 59.3 &  65.43      \\
                                        & \multicolumn{1}{c|}{C-DANN \cite{CDANN}}  & 53.4 & 46.5 & 44.7 & 12.9 & 18.4 & 35.18 & 76.0 & 51.1 & 74.1 & 61.0 &   65.55     \\
                                        & \multicolumn{1}{c|}{EoA \cite{EoA}}  & 65.9 & 57.1 & 55.3 & 16.5 & 23.4 & 43.64 & 81.5 & 59.8 & 79.5 & 69.1 & 72.48        \\
                            & \multicolumn{1}{c|}{SelfReg \cite{kim2021selfreg}}  & 62.4 & 53.7  & 51.7  & 14.7  & 22.5 & 41.0 & 78.8  & 55.4  & 78.4 & 64.9  &  69.37    \\
                            & \multicolumn{1}{c|}{SagNet \cite{sagnets}}  & 57.5 & 49.5  & 46.3  & 13.5  & 19.2 & 37.2 & 78.3 & 54.8 & 75.8  & 63.4 & 68.08       \\
                            & \multicolumn{1}{c|}{ARM \cite{arm}}  & 49.6 & 43.9  & 41.5  & 10.8  & 16.5 & 32.46 & 75.2 & 51.0 & 74.1  & 58.9 & 64.8      \\
                            & \multicolumn{1}{c|}{V-REx \cite{Vrex}}  & 43.3 & 37.7  & 32.5  & 9.8  & 14.1 & 27.48  & 76.6 & 53.0  & 75.3  & 60.7 & 66.4      \\
                            & \multicolumn{1}{c|}{MTL \cite{MTL}}  & 58.0 & 49.0  & 46.2 & 12.7  & 19.2 & 37.02 & 76.8 & 52.4  & 74.9  & 61.5 & 66.4       \\
                            & \multicolumn{1}{c|}{SAND \cite{SAND_mask}}  & 43.8 & 39.9  & 38.2  & 9.0  & 15.2  & 29.22 & 76.2 & 53.3 & 73.5  & 60.3 & 65.82        \\
                            & \multicolumn{1}{c|}{RSC \cite{huangRSC2020}}  & 55.5 & 47.8  & 44.4  & 12.5  & 18.3 & 35.7 & 75.1 & 51.4  & 74.8  & 60.7 & 65.50     \\
                           & \multicolumn{1}{c|}{Fishr \cite{rame2021ishr}}   & 58.3 & 50.5  & 47.9  & 13.6  & 20.2 & 38.1 & 78.3 & 54.4  & 76.2  & 62.4 & 67.83     \\
                           
                           & \multicolumn{1}{c|}{SWAD \cite{cha2021swad}}   & 66.0 & 55.5 & 53.5  & 16.1  &  22.4 & 42.7 & 80.2  & 57.7  & 78.4 & 66.1  &   70.6  \\
   \midrule[0.2mm]
\multirow{1}{*}{ViTs}                  & \multicolumn{1}{c|}{AGG}                                           & 69.14 & 54.25 & 58.15 & 14.83 & 27.55 & 44.78 & 84.64 & 60.10 & 84.43 & 74.2 &    75.84   \\   
                                      
                                      \midrule[0.2mm]
\multirow{3}{*}{MViTs}       & \multicolumn{1}{c|}{Zero-Shot} & 67.8 & 61.79 & 64.13 & 13.9 & \uline{45.7} & 50.66

                        & 84.7 & 60.8 & 83.37 & \uline{78.9} & 76.94  \\ 
                         
                                       & \multicolumn{1}{c|}{Linear Eval} & 63.2 & 59.37 & 57.36 &  10.34 & 41.7 & 46.39 & 82.51 & 66.66 & 81.22 & 72.86  & 75.81 \\ 
                                       & \multicolumn{1}{c|}{Attention Eval} & 75.3 & 64.68 & 64.33 & 16.30 & 44.23 & 52.97 & 88.14 & 69.00 & \uline{88.99} & 77.53 &  80.92 \\ 
                                       \midrule[0.2mm] 
 \multirow{5}{*}{MViTs + ViTs}     & \multicolumn{1}{c|}{Distillation} & 65.23 & 52.29 & 55.55 & 14.06 & 25.8 & 42.59

                        & 85.08 & 59.56 & 83.92 & 74.04 & 75.65  \\   

                                       & \multicolumn{1}{c|}{Cross Attention} & 75.14 &  63.75 & 64.16 & 15.80 & 39.01 & 51.57 & 86.67 & 71.56 & 88.66 & 74.20 &   80.27     \\    
                                       & \multicolumn{1}{c|}{Early Fusion}    & \uline{76.75} & \uline{64.6}  & \uline{65.35} & \uline{17.1}  & 41.86 & \uline{53.13} & \uline{88.76} & 68.86 & 88.33 &  \uline{78.68} &  \uline{81.16} \\ 
                              \rowcolor{mygray}    \cellcolor{white}    &  \multicolumn{1}{c|}{INDIGO}  & \textbf{76.9} & \textbf{65.65} & \textbf{66.42} & \textbf{17.4} & \textbf{46.32} & \textbf{54.54}                       & \textbf{89.38}  & \textbf{73.31} & \textbf{90.78}  & \textbf{79.92}  & \textbf{83.35}  \\ \bottomrule[0.4mm]
\end{tabular}}
\end{center}
\label{table:clip_dg_results}
\vspace{-11mm}
\end{table}
\noindent \textbf{Open Domain Generalization}
Shu et al. \cite{shu2021open} introduce OpenDG, a challenging domain generalization setting where each source domain holds disparate label sets. Since different label sets of distinct source domains cause some classes to be present in more domains than other classes, minor classes' data in a few domains lack diversity. This makes the problem extremely difficult by biasing model representations towards domains than content. Hence, we next evaluate the performance of INDIGO on Office-Home \cite{office-home} and PACS \cite{pacs} datasets under an open setting. \\
\begin{table}[t]
\caption{\scriptsize \textbf{OpenDG results.} Performance of INDIGO on Office-Home (\textbf{R}: real world, \textbf{C}: clipart, \textbf{P}: product, \textbf{A}: art) and PACS (\textbf{P}: photo, \textbf{A}: art, \textbf{C}: cartoon, \textbf{S}: sketch) datasets under open setting. We highlight the \textbf{best results} and the \uline{second best} results. The results are averaged over five runs. INDIGO consistently outperforms all compared methods especially on challenging domains like \textit{sketch}.}
\begin{center}
\setlength{\tabcolsep}{5pt}
\resizebox{\textwidth}{!}{
\begin{tabular}{c|c|c|c|c|c|g|c|c|c|c|g}
\toprule[0.4mm]
\multirow{2}{*}{\textbf{Type}} & \multicolumn{1}{c|}{\multirow{2}{*}{\textbf{Method}}}                     & \multicolumn{5}{c|}{\textbf{Office-Home}}                                                                                                                                                 & \multicolumn{5}{c}{\textbf{PACS}}                                                                                                             \\ 
                                       & \multicolumn{1}{c|}{}                                                     & \multicolumn{1}{c|}{\textbf{R}} & \multicolumn{1}{c|}{\textbf{C}} & \multicolumn{1}{c|}{\textbf{P}} & \multicolumn{1}{c|}{\textbf{A}}  & \textbf{Avg.} & \multicolumn{1}{c|}{\textbf{P}} & \multicolumn{1}{c|}{\textbf{A}} & \multicolumn{1}{c|}{\textbf{C}} & \multicolumn{1}{c|}{\textbf{S}} & \textbf{Avg.} \\ \midrule
\multirow{2}{*}{CNNs}                  & \multicolumn{1}{c|}{AGG} & 62.4 & 42.83 & 54.27 & 42.22 & 50.43 & 53.15 & 51.35 & 66.43 & 49.75 &    55.17          \\
                                        & \multicolumn{1}{c|}{MLDG \cite{metadg1}} & 62.98  & 41.82 & 56.89 & 42.58 & 51.07 & 62.20 & 44.59 & 71.64 &  51.29 &     45.00         \\
                                        & \multicolumn{1}{c|}{FC \cite{Li2019ICML_FC}} & 63.79 & 41.80 &  54.41 & 44.13 & 51.03 & 60.94 & 51.12 &  69.32 & 51.15 &   58.13           \\
                                        & \multicolumn{1}{c|}{Epi-FCR \cite{li2019episodic}} &  62.60 & 37.13 & 54.95 & 46.33 & 50.25 &  46.35 & 54.16 & 72.00 & 46.35 &    60.64           \\
                                        & \multicolumn{1}{c|}{PAR \cite{PAR_dg}} &  65.98 & 41.27 &  55.37 &  42.40 & 51.26 & 51.86 & 52.97 & 62.77 & 53.62 & 56.56              \\
                                        & \multicolumn{1}{c|}{RSC \cite{huangRSC2020}} & 60.85 & 38.60 &  54.61 &  44.19 & 49.56 &  67.53 & 50.47 & 67.51 & 50.17 & 58.92              \\
                                        & \multicolumn{1}{c|}{CuMix \cite{mancini2020towards}} &  64.63  & 41.54 & 57.74 &  42.76 &  51.67 & 65.67 & 53.85 & 74.16 & 37.70 & 57.85              \\
                                        & \multicolumn{1}{c|}{DAML \cite{shu2021open}} & 65.99 & 45.13 &  61.54 &  53.13 &  56.45 & 75.69 & 43.02 & 73.65 & 58.50 & 65.49              \\
                                        \midrule[0.2mm]
\multirow{1}{*}{ViTs}  
                                      & \multicolumn{1}{c|}{AGG}                                            & 76.71  & 53.76& 67.39 & 65.35 & 65.80 & 59.55 & 63.70 & 52.15 & 34.12  &  52.38            \\
                                      \midrule[0.2mm]
\multirow{5}{*}{MViTs}       & \multicolumn{1}{c|}{Zero-Shot} & \uline{81.2} & 63.69 & \uline{82.33} & \textbf{70.8} & \uline{74.5} & \textbf{99.99} & \textbf{97.87} &  \textbf{99.53} & \uline{87.34} & \textbf{96.18} \\ 
                                       & \multicolumn{1}{c|}{Linear Eval}     & 52.97 & 48.07 & 50.32 & 47.65 & 49.75 & 79.26 & 82.71 & 82.29 & 72.76 &   79.26           \\ 
                                       & \multicolumn{1}{c|}{Attention Eval}   & 75.41 & \uline{72.75} & 62.83 & 63.08 & 68.52 & 76.92 & 78.23 & 80.48 & 78.17 & 78.45 \\  
                                       \midrule[0.2mm] 
                                       
\multirow{5}{*}{MViTs + ViTs}       & \multicolumn{1}{c|}{Distillation}  & 79.52 & 63.69 & 56.37 & 67.08 & 66.66 & 65.10 & 59.16 &  53.72 & 38.52 & 54.12 \\
                                    & \multicolumn{1}{c|}{Early Fusion} & 75.77 & 59.03 & 70.44 & 65.89 & 67.78 & 77.35 & 68.59 & 74.67 & 61.89 &    70.62          \\
                                       & \multicolumn{1}{c|}{Cross attention}  &  76.13 & 67.86 & 74.27 & 64.22 & 70.62 & 76.52 & 77.44 & 87.88 & 83.29 &   81.28           \\
                                       
                      \rowcolor{mygray} \cellcolor{white}                 & \multicolumn{1}{c|}{INDIGO} & \textbf{83.23} & \textbf{73.25} & \textbf{83.51} & \uline{67.68} & \textbf{76.91} & \uline{93.44}  & \uline{93.61}  & \uline{91.08} & \textbf{90.45}  & \uline{92.14} \\ \bottomrule[0.4mm]
\end{tabular}}
\end{center}
\label{table:clip_opendg_results}
\vspace{-8mm}
\end{table}
\noindent \textit{(\textbf{Baselines})} Similar to previous setting, we compare all four kinds of pipelines - (1) CNNs, which includes prior arts and current state-of-the-art, DAML \cite{shu2021open} that uses three Resnet-18 backbones (comparable to Resnet-50); (2) ViTs, which include DeiT-S \cite{deit} trained in AGG manner; (3) MViTs, which include conventional ways of using the pre-trained MViT; and (4) MViTs + ViTs, that include our newly introduced fusion baselines and our proposed fusion, INDIGO (all using a DeiT-S visual backbone). Implementation and architectural details of fusion module are described in the supplementary.\\
\noindent \textit{(\textbf{Training and evaluation protocol})}  Similar to DAML \cite{shu2021open}, we consider each domain as the target domain and the rest domains as source domains for training. We use training-domain validation model selection criteria for both datasets. We report the accuracy of target domain data from non-open classes as in \cite{shu2021open}. \\
\noindent \textit{(\textbf{Results})} Table \ref{table:clip_opendg_results} presents our results when CLIP-ViT-B/16 \cite{Radford2021LearningTV} is used as an MViT in the multimodal branch. It can be seen that even in challenging settings like OpenDG, where there is a high chance of model representations becoming domain biased, INDIGO achieves state-of-the-art results on the Office-Home dataset. On PACS, even though zero-shot inference works best on average, INDIGO still performs best on challenging \textit{sketch} domain (on which all other methods perform worse). Since PACS (under open setting) is a relatively smaller and less-complex (having only six non-open classes) dataset than Office-Home (having 54 non-open classes), we believe it led to overfitting/memorization of the source domain data. This can also be seen with ViTs (trained with vanilla loss), which significantly outperforms state-of-the-art approach DAML \cite{shu2021open} on Office-Home but overfits on PACS. \\
\noindent \textbf{Choice of MViT.} Apart from CLIP, we also experiment with two other pre-trained MViTs - DeCLIP \cite{li2021supervision} and ALBEF \cite{ALBEF}. DeCLIP uses additional self, multi-view, and nearest-neighbor supervision along with image-text contrastive supervision to achieve similar performance as CLIP but with 7.1 x fewer data. ALBEF, on the other hand, uses momentum distillation, a self-training method to learn from pseudo-targets produced by a momentum model. As shown in Table \ref{table:different_multimodal_networks}, INDIGO still outperforms conventional and our newly introduced baselines with good margins on the Office-Home dataset under both closed and open settings. This highlights the efficacy of INDIGO when other pre-trained multimodal networks are used in the multimodal branch. Overall, CLIP, when used in INDIGO, performs best. \\
\begin{table}[t]
\caption{\scriptsize \textbf{Results with other MViTs.} Performance of INDIGO with different MViTs on Office-Home (\textbf{R}: real world, \textbf{C}: clipart, \textbf{P}: product, \textbf{A}: art) under closed and open setting. We highlight the \textbf{best results} and the \uline{second best} results. The results are averaged over five runs. INDIGO consistently achieves best results compared to conventional and newly introduced baseline approaches.}
\begin{center}
\setlength{\tabcolsep}{5pt}
\resizebox{\textwidth}{!}{
\begin{tabular}{c|c|c|c|c|c|g|c|c|c|c|g}
\toprule[0.4mm]
\multirow{2}{*}{\textbf{Multimodal}} & \multicolumn{1}{c|}{\multirow{2}{*}{\textbf{Method}}}                     & \multicolumn{5}{c|}{\textbf{Closed Office-Home}}                                                                                                                                                 & \multicolumn{5}{c}{\textbf{Open Office-Home}}                                                                                                             \\ 
                                       & \multicolumn{1}{c|}{}                                                     & \multicolumn{1}{c|}{\textbf{R}} & \multicolumn{1}{c|}{\textbf{C}} & \multicolumn{1}{c|}{\textbf{P}} & \multicolumn{1}{c|}{\textbf{A}}  & \textbf{Avg.} & \multicolumn{1}{c|}{\textbf{R}} & \multicolumn{1}{c|}{\textbf{C}} & \multicolumn{1}{c|}{\textbf{P}} & \multicolumn{1}{c|}{\textbf{A}} & \textbf{Avg.} \\ \midrule
\multirow{6}{*}{CLIP}       & \multicolumn{1}{c|}{Zero-Shot}      &       84.7 & 60.8 & 83.37 & \uline{78.9} & 76.94 &  \uline{81.2} & 63.69 & \uline{82.33} & \textbf{70.8} & \uline{74.5} \\
                                       & \multicolumn{1}{c|}{Linear Eval} &  82.51 & 66.66 & 81.22 & 72.86  & 75.81                  & 52.97 & 48.07 & 50.32 & 47.65 & 49.75   \\  
                                       & \multicolumn{1}{c|}{Attention Eval}  & 88.14 & 69.00 & \uline{88.99} & 77.53 &  80.92  & 75.41 & \uline{72.75} & 62.83 & 63.08 & 68.52                      \\ 
                                        
                                       & \multicolumn{1}{c|}{Cross-attention} & 86.67 & \uline{71.56} & 88.66 & 74.20 &   80.27  & 76.13 & 67.86 & 74.27 & 64.22 & 70.62 \\  
                                       & \multicolumn{1}{c|}{Early Fusion} &  \uline{88.76} & 68.86 & 88.33 &  78.68 &  \uline{81.16} & 75.77 & 59.03 & 70.44 & 65.89 & 67.78 \\ 
                                 \rowcolor{mygray} \cellcolor{white}        & \multicolumn{1}{c|}{INDIGO} & \textbf{89.38}  & \textbf{73.31} & \textbf{90.78}  & \textbf{79.92}  & \textbf{83.35} & \textbf{83.23} & \textbf{73.25} & \textbf{83.51} & \uline{67.68} & \textbf{76.91} \\ \midrule[0.3mm]
\multirow{5}{*}{DeCLIP}       &  \multicolumn{1}{c|}{Linear Eval} &  34.74 & 37.92 & 41.49 & 33.1 & 36.8 & 15.37 & 8.48 & 14.87 & 8.28 & 11.75 \\ 
                                       & \multicolumn{1}{c|}{Attention Eval}  & 86.36  & \uline{70.53} & \uline{88.45} & 70.16 & 78.87 & \uline{74.51} & \uline{66.69} & \uline{70.55} & \uline{60.10} & \uline{67.96}\\
                                       &  \multicolumn{1}{c|}{Cross-attention} & 79.48 & 65.14 & 81.34 & 63.20 & 72.29 & 56.68 & 55.57 & 54.18 & 46.57 &    53.25        \\  
                                       & \multicolumn{1}{c|}{Early Fusion}    & \uline{87.68} & 67.23 & 88.41 & \textbf{73.57} & \uline{79.22} & 70.51 & 59.65 & 67.90 & 59.92 & 64.50  \\ 
                             \rowcolor{mygray} \cellcolor{white}            & \multicolumn{1}{c|}{INDIGO}    & \textbf{88.61} & \textbf{73.28} & \textbf{90.45} & \uline{73.05} & \textbf{81.35} & \textbf{83.17} & \textbf{69.70} & \textbf{76.41} & \textbf{62.61} &  \textbf{72.97}                                   \\ \midrule[0.3mm]
\multirow{5}{*}{ALBEF}       & \multicolumn{1}{c|}{Linear Eval}  & 84.20 & 70.74 & 83.19 & 77.32 & 78.86 & 69.70 & 60.04 & 65.31 & 64.04 & 64.77 \\ 
                                       & \multicolumn{1}{c|}{Attention Eval}   & 85.86 & 69.40 & 86.30 & 75.85 & 79.35 & 74.61 & \uline{66.45} & 71.68 & 62.0 & 68.68 \\ 
                                       & \multicolumn{1}{c|}{Cross-attention}  & 86.29 & \uline{71.15} & 86.52 & 73.04 & 71.75 & 74.64 & 65.71 & 67.5 & 62.61 &    67.61        \\  
                                       & \multicolumn{1}{c|}{Early Fusion}    &  \uline{87.33} & 69.49 & \uline{87.27} & \uline{77.77} & \uline{80.46} & \uline{76.55} & 64.03 & \uline{73.11} & \uline{66.48} & \uline{70.04}  \\ 
                            \rowcolor{mygray} \cellcolor{white}             & \multicolumn{1}{c|}{INDIGO}     & \textbf{87.52} & \textbf{73.42} & \textbf{87.46} & \textbf{78.68} & \textbf{81.77} & \textbf{82.59} & \textbf{71.47} & \textbf{77.79} & \textbf{68.39} &  \textbf{75.06}                                  \\ \bottomrule[0.4mm]
\end{tabular}}
\end{center}
\label{table:different_multimodal_networks}
\vspace{-10mm}
\end{table}
\noindent \textbf{Choice of visual network and number of layers in fusion module.} To highlight that INDIGO is leveraging the visual modality, we perform an ablation where we vary the strength of ViT used in the visual branch. Additionally, we also vary the number of layers used in the fusion module to show its effect on final performance. As shown in Table \ref{table:varying_vit}, by using more powerful (Hybrid ViTs) and large (ViT-B) vision transformers \cite{vit} in the visual branch, the domain generalization performance of INDIGO improves. This shows that INDIGO can attend to visual modality to learn additional shape-biased concepts, and the performance is not solely because of intrinsic modality. The gain in performance becomes prominent when more layers are used in the fusion module, implying a better inter-modality interaction between intrinsic and visual modality tokens. 
\begin{table}[]
\vspace{-7mm}
\begin{center}
\caption{\scriptsize \textbf{Ablation on choice of visual network and number of layers in fusion module.} Performance of INDIGO when different networks are used in visual branch and layers of fusion module are increased on Office-Home (\textbf{R}: real world, \textbf{C}: clipart, \textbf{P}: product, \textbf{A}: art) under closed setting. The results are averaged over five runs. Stronger and Larger ViTs can be seen to further improve the generalization of INDIGO to unseen domains. } 
\vspace{10pt}
\setlength{\tabcolsep}{5pt}
\resizebox{0.9\textwidth}{!}{
\begin{tabular}{c|c|c|c|c|g|c|c|c|c|g}
\toprule[0.4mm]
\multirow{2}{*}{\textbf{Backbone}} & \multicolumn{5}{c|}{\textbf{3 Layers}}                                                                                                                                                 & \multicolumn{5}{c}{\textbf{12 Layers}}                                                                                                             \\ 
                                       & \multicolumn{1}{c|}{\textbf{R}} & \multicolumn{1}{c|}{\textbf{C}} & \multicolumn{1}{c|}{\textbf{P}} & \multicolumn{1}{c|}{\textbf{A}}  & \textbf{Avg.} & \multicolumn{1}{c|}{\textbf{R}} & \multicolumn{1}{c|}{\textbf{C}} & \multicolumn{1}{c|}{\textbf{P}} & \multicolumn{1}{c|}{\textbf{A}} & \textbf{Avg.} \\ \midrule
Resnet-50 & 88.8 & 72.91 & 90.1 & 79.34 & 82.78 & 89.0 & 72.48 &  90.2 & 78.2 & 82.47 \\
DeiT-Ti  & 89.02 & 72.88 & 90.14 & 80.05 & 83.02 & 89.34 & 73.52 & 90.43 & 79.31 & 83.15 \\
Hybrid-ViT-Ti & 89.1 & 72.77 & 90.3 & 79.94 & 83.02 &  89.7 & 73.82 & 90.50 & 79.9 & 83.48 \\
DeiT-S  & 89.38 & 73.31 & 90.78 & 79.92 & 83.35 & 90.1 & 74.32 & 90.99  & 80.2 & 83.90 \\
Hybrid-ViT-S & 89.73 & 73.71 & 91.05 & 81.16 & 83.91 & 90.88  & 75.45 & 91.22 & 81.62 & 84.80 \\
ViT-B & 91.4 & 74.23 & 91.84 & 82.33 & 84.95 & 91.76 & 75.85 & 92.13 & 83.51 & 85.81 \\\bottomrule[0.4mm]
\end{tabular}}
\label{table:varying_vit}
\end{center}
\vspace{-10mm}
\end{table}

\noindent \textbf{Choice of fusion mechanism.} We analyze how different fusion mechanisms perform as compared to multi-head self-attention modules (MSA) \cite{vit}, which we currently use in our fusion module. In Table \ref{table:fusion_mechanisms}, we compare - (1) naive fusion mechanism like concatenation (which just concatenates the intrinsic and visual modality tokens), (2) multi-head self-attention (MSA) \cite{vit}, (3) multi-head cross-attention (MCA) \cite{chen2021crossvit}, and (4) MLP-mixer \cite{tolstikhin2021mlpmixer}. We observe that MCA performs slightly better than MSA, whereas MLP-mixer and concatenation perform inferior. This shows that the choice of fusion mechanism can affect the overall performance, which we leave for future works to explore.\\
\noindent \textbf{Can fine-tuning MViT help further? } In all our previous experiments, we used a frozen pre-trained multimodal network. As an additional experiment, along with training the visual branch and fusion module, we also finetune the multimodal network (i.e CLIP) . We finetune in two ways - (1) only normalization layers; and (2) last layer. Table \ref{table:finetuning} shows that the performance of INDIGO further improves while still outperforming standalone finetuning of CLIP.\par
\begin{figure}
\begin{minipage}{0.65\textwidth}
\centering
    \includegraphics[width=\textwidth]{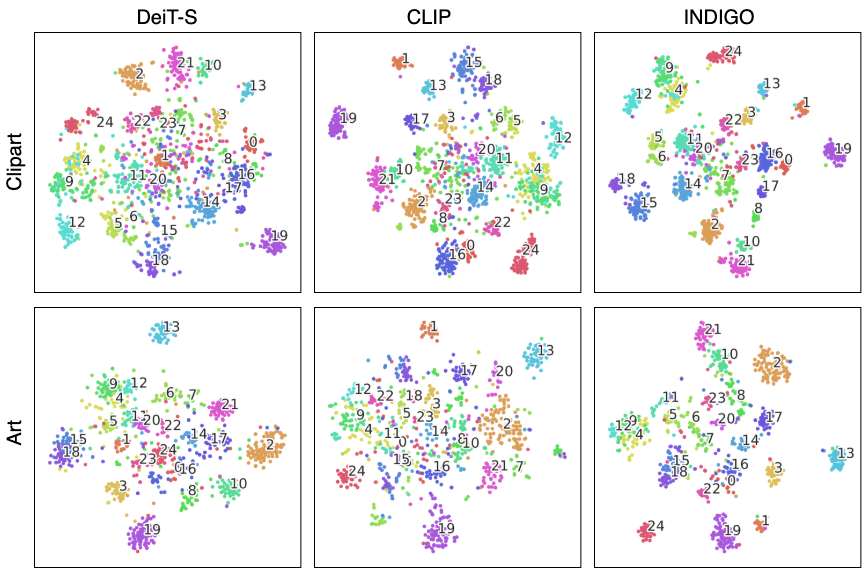}
    \subcaption{}
    \label{fig:tsne_plots}
\end{minipage}
\hfill
\begin{minipage}{0.34\textwidth}
\centering
    \includegraphics[width=\textwidth]{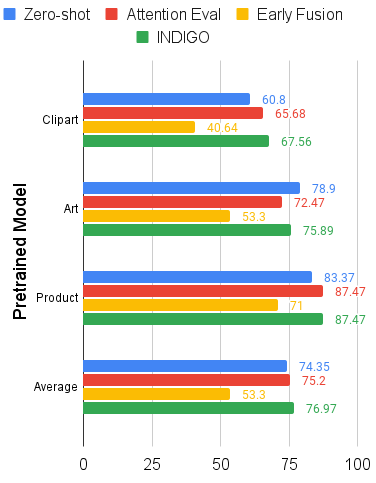}
    \subcaption{}
    \label{fig:limited_sources}
\end{minipage}
\caption{\scriptsize (a) \textbf{t-SNE plots.} t-SNE visualization of learned feature representations by DeiT-S (standard AGG training), CLIP and our proposed INDIGO method when \textit{clipart} and \textit{art} are chosen as target domains for Office-Home dataset, (b) \textbf{Limited sources DG.} Performance of INDGIO when trained only on \textit{real world} as source domain and evaluated on \emph{clipart}, \emph{art} and \emph{product} as unseen target domains for Office-Home dataset. \textit{(Best viewed in color, zoomed in)} }
\vspace{-7.5mm}
\end{figure}
\begin{figure}
    \centering
    \includegraphics[width=\textwidth]{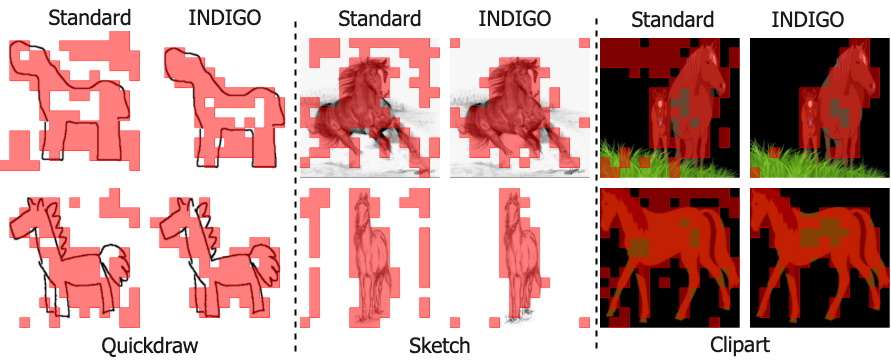}
    \caption{\textbf{Attention maps} for DeiT-S on harder \emph{quickdraw}, \emph{sketch} domains and relatively simpler \emph{clipart} domain trained with \emph{Standard} (AGG) training and using our proposed \emph{INDIGO} method. Best attention heads depicted for both approaches.}
    \label{fig:attention_maps}
\vspace{-10mm}
\end{figure}

\vspace{-20pt}
\begin{table}[]
\begin{center}
\begin{minipage}{0.49\textwidth}
\caption{\scriptsize \textbf{Ablation on finetuning the MViT.} Performance of INDIGO when MViT is also finetuned on Office-Home (\textbf{R}: real world, \textbf{C}: clipart, \textbf{P}: product, \textbf{A}: art) under closed setting. We use training-domain validation set model selection criteria. The results are averaged over five runs.}
\vspace{10pt}
\setlength{\tabcolsep}{5pt}
\resizebox{\textwidth}{!}{
\begin{tabular}{c|c|c|c|c|g}
\toprule[0.4mm]
\multirow{2}{*}{\textbf{Backbone}} & \multicolumn{5}{c}{\textbf{Closed OfficeHome}}                                                                                                       \\ 
                                       & \multicolumn{1}{c|}{\textbf{R}} & \multicolumn{1}{c|}{\textbf{C}} & \multicolumn{1}{c|}{\textbf{P}} & \multicolumn{1}{c|}{\textbf{A}}  & \textbf{Avg.}\\  \midrule
CLIP (FT B.N Layers) & 84.00 & 68.92 & 84.13 & 76.47 & 78.38\\
\rowcolor{mygray}  INDIGO (FT B.N Layers) & 89.50 & 73.50 & 91.02 & 80.2 & 83.55 \\
\midrule
CLIP (FT Last Layers) & 89.6 & 74.5 & 89.70 & 83.43 & 84.30 \\
\rowcolor{mygray}  INDIGO (FT Last Layer) & 90.83 & 76.87 & 91.64 & 83.91 & 85.81 \\
\bottomrule[0.4mm]
\end{tabular}}
\label{table:finetuning}
\end{minipage}
\hfill
\begin{minipage}{0.49\textwidth}
\caption{\scriptsize \textbf{Ablation on choice of fusion mechanism.} Performance of INDIGO when different fusion mechanisms are used in fusion module on Office-Home (\textbf{R}: real world, \textbf{C}: clipart, \textbf{P}: product, \textbf{A}: art) under closed setting. The results are averaged over five runs.} 
\vspace{10pt}
\setlength{\tabcolsep}{5pt}
\resizebox{\textwidth}{!}{
\begin{tabular}{c|c|c|c|c|g}
\toprule[0.4mm]
\multirow{2}{*}{\textbf{Backbone}} & \multicolumn{5}{c}{\textbf{Closed OfficeHome}}                                                                                                       \\ 
                                       & \multicolumn{1}{c|}{\textbf{R}} & \multicolumn{1}{c|}{\textbf{C}} & \multicolumn{1}{c|}{\textbf{P}} & \multicolumn{1}{c|}{\textbf{A}}  & \textbf{Avg.}\\  \midrule
Concatenation & 83.37 & 60.78 & 84.79 & 75.06 & 76.0 \\
MSA & 89.38 & 73.31 & 90.78 & 79.92 & 83.35\\
MCA &  89.87 & 73.30 & 90.71 & 80.03 & 83.47 \\
MLP-Mixer & 88.93 & 72.52 & 90.23 & 78.75 & 82.60 \\
\bottomrule[0.4mm]
\end{tabular}}
\label{table:fusion_mechanisms}
\end{minipage}
\end{center}
\end{table}
\vspace{-20pt}
\noindent \textbf{t-SNE plots.} We analyze and compare the representations learned by INDIGO with DeiT-S \cite{deit} and CLIP \cite{Radford2021LearningTV} on target domain (for 25 classes of Office-Home) via t-SNE plots in Figure \ref{fig:tsne_plots}. As can be seen, for INDIGO, the plot is less noisy and well segregated into class clusters as compared to DeiT-S and CLIP, resulting in state-of-the-art generalization on these target domains.\\
\noindent \textbf{DG with limited sources.} Generalization to unseen domains can become challenging when data from only a few source domains is available. To test INDIGO under such a challenging scenario, we experiment on Office-Home with \textit{real world} as the only source domain and rest domains (\textit{real world}, \textit{clipart}, \textit{product}) as target domains. In Figure \ref{fig:limited_sources}, we can observe that INDIGO still results in the best average performance when compared with zero-shot CLIP, attention eval (on frozen CLIP features), and Early Fusion baselines.  \\
\noindent \textbf{Attention maps.} Similar to \cite{naseer2021intriguing}, in Figure \ref{fig:attention_maps}, we analyze and compare attention maps of a DeiT-S trained in vanilla (AGG) fashion with the one used in INDIGO's visual branch. We observe that, when used in INDIGO pipeline, the vision transformer can concentrate on foreground objects in the scene and better ignore the background or style. This confirms that attending with intrinsic modality in the fusion module helps the visual transformer exhibit more shape-bias than a vanilla one. The experiment is performed on DomainNet with \textit{quickdraw}, \textit{sketch} and \textit{clipart} as target domains separately.
\vspace{-10pt}
\section{Conclusions}
\vspace{-8pt}
In this work, we study how multimodal information present in pre-trained vision-language models can be leveraged ``intrinsically" to build systems that generalize to unseen domains. We propose INDIGO, a simple and elegant way to combine the intrinsic  and  visual  modalities  obtained  from  pre-trained  multimodal network and vision transformer (ViT), respectively. We conduct extensive experiments to demonstrate the effectiveness of the approach in generalizing to unseen domains under closed, open, and limited sources settings. We then conduct a thorough analysis to characterize the efficacy of our approach in leveraging both intrinsic and visual  modalities. Our future work will include the development of better methods to effectively fuse both modalities to improve generalization performance in unseen domains further. We also plan to extend and explore the significance of our work in other challenging settings like OOD generalization, data-free domain generalization, zero-shot domain generalization,  domain generalized semantic segmentation, and visual grounding.

%
%
\bibliographystyle{splncs04}
\bibliography{egbib}
\end{document}


\pagestyle{headings}
\mainmatter
\renewcommand{\thesection}{S\arabic{section}}
\renewcommand{\thefigure}{S\arabic{figure}}
\renewcommand{\thetable}{S\arabic{table}}
\def\ECCVSubNumber{7762}  

\title{Supplementary Section: \\ \textit{INDIGO}: Intrinsic Multimodality for Domain Generalization} 

\titlerunning{INDIGO}
%
\author{Puneet Mangla \inst{1}\thanks{Equal contribution.}\and
Shivam Chandhok \inst{3}* \and
Milan Aggarwal\inst{1} \and
Vineeth N Balasubramanian \inst{2} \and
Balaji Krishnamurthy \inst{1}
}
\authorrunning{Puneet, Shivam et al.}
%
\institute{Adobe Media and Data Science Research Lab,
Noida, India 
\and Indian Institute of Technology, Hyderabad
\and INRIA, Universite Grenoble Alpes  \\
\email{\{pmangla261, chandhokshivam, milan.ag1994\}@gmail.com, vineethnb@iith.ac.in, kbalaji@adobe.com} \\}

\maketitle
\appendix
In this supplementary section, we discuss the following details, which could not be included in the main paper owing to space constraints:
\vspace{-6pt}
\begin{itemize}[leftmargin=*]
\setlength\itemsep{-0.03em}
    \item[\ref{section:analysis}] Additional analysis for the proposed INDIGO approach on: 
    \begin{itemize}
        \item Choice of CLIP architecture for multimodal branch.
        \item Results on Limited data DG.
        \item Additional t-SNE visualizations.
        \item Additional attention maps.
    \end{itemize}
   \item[\ref{section:implementation}] Implementation details of INDIGO to facilitate reproducibility, including:
    \begin{itemize}
        \item Architectural details.
        \item Training details.
        \item Effect of Varying hyperparameter $\lambda$.
    \end{itemize}
\end{itemize}
\section{Additional Analysis on Office-Home} \label{section:analysis}
\begin{figure}[h]
    \centering
    \includegraphics[width=\textwidth]{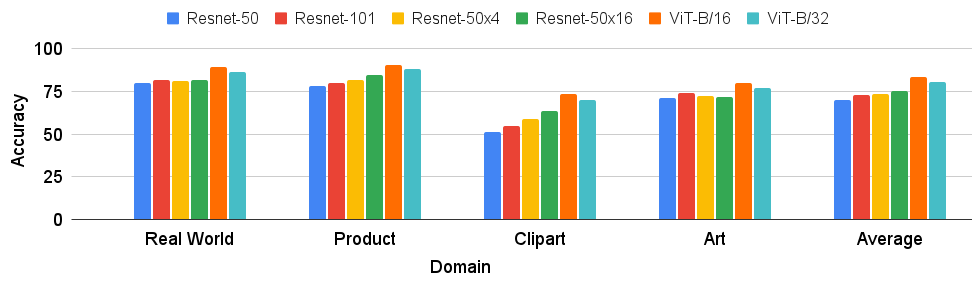}
    \caption{\textbf{Choice of CLIP architecture.} Performance of INDGIO when different CLIP architectures are used in multimodal branch.}
    \label{fig:clip_architecture}
\vspace{-6mm}
\end{figure}

\noindent \textbf{Choice of CLIP architecture.} In our main results (Table 1 and 2 of main manuscript), we used pre-trained CLIP-ViT-B/16 network [61] for the multimodal branch. In order to validate this choice, here we show the performance of our approach when other pre-trained CLIP architectures such as Resnet-50, Resnet-101, Resnet-50x4, Resnet-50x16, and ViT-B/32 are used in the multimodal branch. Figure \ref{fig:clip_architecture} shows that using powerful architectures like Resnet-50x4, Resnet-50x16 (in comparison to Resnet-50 and Resnet-101) improves generalization on challenging domains like \textit{clipart} and \textit{product}. However, we observe that the attention based vision transformer architectures consistently outperform the convolutional counterparts for all domains. Further, we observe that using a vision transformer with large patch size (ViT-B/32) yields inferior performance compared to one using a smaller patch size (ViT-B/16).\\
\noindent \textbf{Limited Data DG.} We further analyze how INDIGO performs in comparison to state-of-art methods, conventional and (proposed) novel baselines when 50\%, 75\%, and 100\% of the original training data is used for training. For this analysis, we randomly choose $x \in \{50, 75, 100 \} \%$ of samples from all source domains and use it for training the models. We repeat the process five times and report the average performance as shown in Table \ref{table:limited_data}. We observe that even with just 50\% of training data, INDIGO outperforms CNN based state-of-the-arts EoA [2] and SWAD [6] (using 100\% of available data). On challenging domains like \textit{clipart}, INDIGO trained with just 50\% of training data beats conventional (Zero-shot, Attention Eval) and novel baseline approaches (Early Fusion). INDIGO trained with 75\% of training data on average outperforms all compared methods (trained on full data).
\begin{table}[t]
\centering
\caption{\textbf{Results on Limited data DG.} Performance of INDIGO when 50\%, 75\%  and 100 \% of the original training data is used for training on Office-Home (\textbf{R}: real world, \textbf{C}: clipart, \textbf{P}: product, \textbf{A}: art) under closed setting. We use training-domain validation set model selection criteria. The results are averaged over five runs.}
\vspace{10pt}
\setlength{\tabcolsep}{5pt}
\resizebox{0.8\textwidth}{!}{
\begin{tabular}{c|c|c|c|c|c|g}
\toprule[0.4mm]
\textbf{Method} & \textbf{Data} & \multicolumn{5}{c}{\textbf{Closed OfficeHome}}   \\ 
 &  & \multicolumn{1}{c|}{\textbf{R}} & \multicolumn{1}{c|}{\textbf{C}} & \multicolumn{1}{c|}{\textbf{P}} & \multicolumn{1}{c|}{\textbf{A}}  & \textbf{Avg.}\\  \midrule
SWAD & 100 \% &  80.2  & 57.7  & 78.4 & 66.1  &   70.6 \\ \midrule
EoA & 100 \% & 81.5 & 59.8 & 79.5 & 69.1 & 72.48  \\ \midrule
Zero-shot & - & 84.7 & 60.8 & 83.37 & 78.9 & 76.94 \\ \midrule
\multirow{4}{*}{Attention Eval} 
& 50 \% & 86.52 & 68.32  &  88.28 & 74.97 & 79.50\\
& 75 \% & 87.78 & 68.55 & 88.73 & 77.01 & 80.51 \\
& 100 \% & 88.14 & 69.00 & 88.99 & 77.53 &  80.92 \\ \midrule
\multirow{4}{*}{Early Fusion} 
 &  50 \% & 67.97 & 44.79 &  71.60 & 57.12 & 60.37\\
  &  75 \% & 73.25 & 45.22 & 68.15  & 59.29 & 61.47 \\
   &  100 \% & 88.76 & 68.86 & 88.33 &  78.68 &  81.16 \\ \midrule
\multirow{4}{*}{INDIGO} 
 &  50 \% & 87.14 & 70.43 & 88.31 & 77.09 & 80.74 \\
  &  75 \% & 88.92 & 72.59 & 90.36 & 79.50 & 82.84\\
   &  100 \% & 89.38 & 73.31 & 90.78  & 79.92  & 83.35  \\
\bottomrule[0.4mm]
\end{tabular}}
\label{table:limited_data}
\vspace{-6mm}
\end{table}

\begin{figure}[t]
    \centering
    \includegraphics[width=\textwidth]{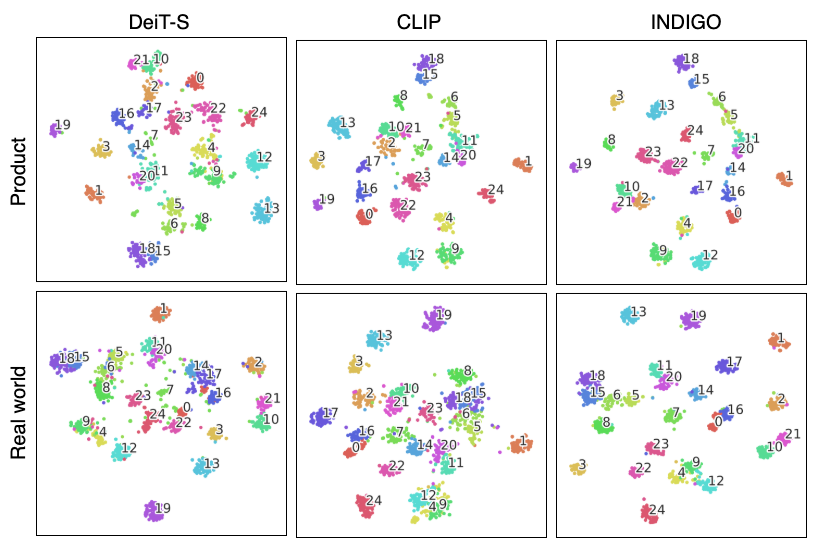}
    \caption{\textbf{t-SNE plots on Office-Home (ClosedDG).} t-SNE visualization of learned feature representations of DeiT-S (standard AGG training), CLIP and our proposed INDIGO method when \textit{product} and \textit{real world} are chosen as target domains for Office-Home dataset (under closed setting).}
    \label{fig:tsne_plots2}
    \vspace{-5mm}
\end{figure}

\begin{figure}
    \centering
    \includegraphics[width=\textwidth]{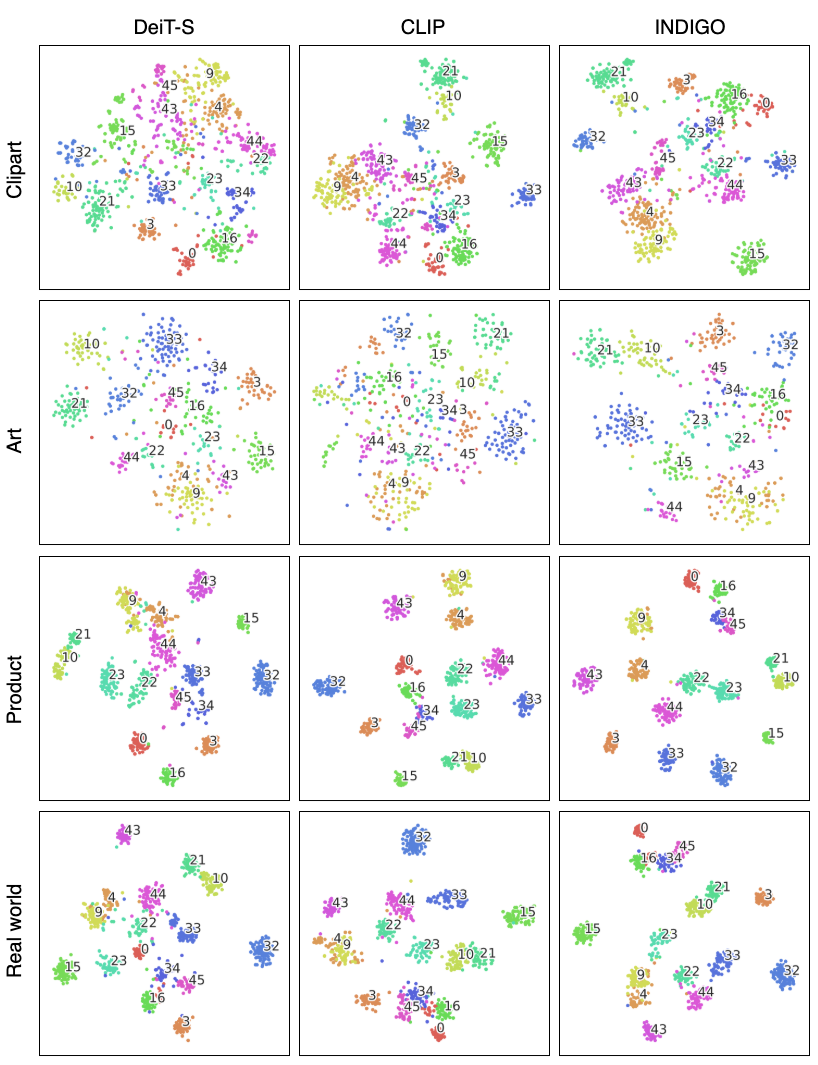}
    \caption{\textbf{t-SNE plots on Office-Home (OpenDG).} t-SNE visualization of learned feature representations of DeiT-S (standard AGG training), CLIP and our proposed INDIGO method when \textit{clipart}, \textit{art}, \textit{product} and \textit{real world} are chosen as target domains for Office-Home dataset (under open setting).}
    \label{fig:tsne_plots_open}
\end{figure}

\noindent \textbf{Additional t-SNE plots.} In Figures \ref{fig:tsne_plots2} and \ref{fig:tsne_plots_open}, we show additional t-SNE plots comparing the representations learned by INDIGO, DeiT-S [76], and CLIP [61] on target domains of Office-Home under (closed) DG (for 25 classes) and OpenDG (for 16 classes) setting, respectively. It can be clearly seen that for our proposed INDIGO approach, the plots  are well segregated into class clusters and are more compact as compared to DeiT-S and CLIP where the class datapoints are potentially leak into other class clusters, resulting in mis-classifications. Thus our method shows better generalization performance on these target domains.\\

\begin{figure}[t]
    \centering
    \includegraphics[width=\textwidth]{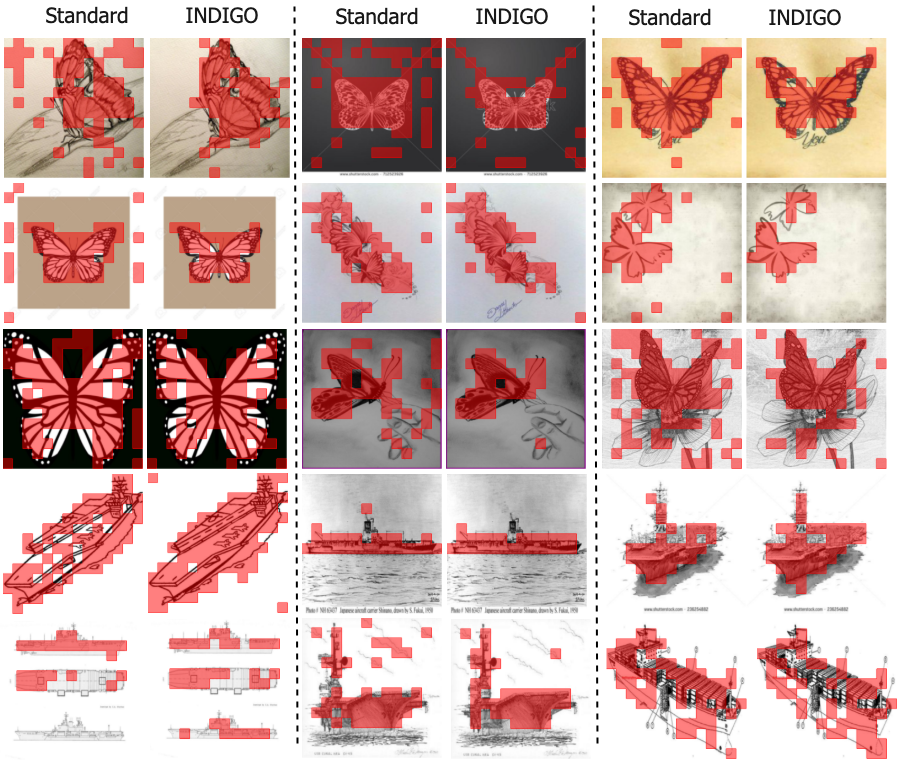}
    \caption{\textbf{Attention maps} for DeiT-S trained with \emph{Standard} (AGG) training and using our proposed \emph{INDIGO} method. Target domain is \emph{sketch}. Best attention heads depicted for both approaches.}
    \label{fig:attention_maps2}
    \vspace{-5mm}
\end{figure}

\noindent \textbf{Additional attention maps.} In Figure \ref{fig:attention_maps2}, we provide additional attention maps of a DeiT-S trained in vanilla (AGG) fashion and the one used in INDIGO's visual branch. Similar to Figure 5 of main submission, we observe that, when used in INDIGO pipeline, the vision transformer can concentrate on foreground objects in the scene and better ignore the background or style. This confirms that attending with intrinsic modality in the fusion module helps the visual transformer exhibit more shape-bias than a vanilla one. The experiment is performed on DomainNet with  \textit{sketch} as target domain.
\section{Implementation Details}
\label{section:implementation}
\noindent \textbf{Architectural details.} For both DomainNet and Office-Home datasets in ClosedDG setting, the fusion module of INDIGO is composed of three layers, each of which includes \texttt{MSA} (with six heads), \texttt{FFN} and \texttt{LN} modules. In case of OpenDG, the same is composed of 5 and 12 layers of six-headed \texttt{MSA} modules for Office-Home and PACS respectively. As mentioned in main submission, we use DeiT-S [76] (considered equivalent to Resnet-50) as $f^V(.)$ in our visual branch to handle visual modality. Linear projections $w^{M}(.)$ and $w^V(.)$ project the intrinsic and visual modality to $384$-dimensional tokens. In OpenDG, the semantic projection layers $p^M(.)$ and $p^V(.)$ projects $\mathbf{x}_K^{M}$ and $\mathbf{x}_K^{V}$ to a  $512$-dimensional semantic space. \par
\begin{figure}[b]
    \centering
    \vspace{-25mm}
    \includegraphics[width=\textwidth]{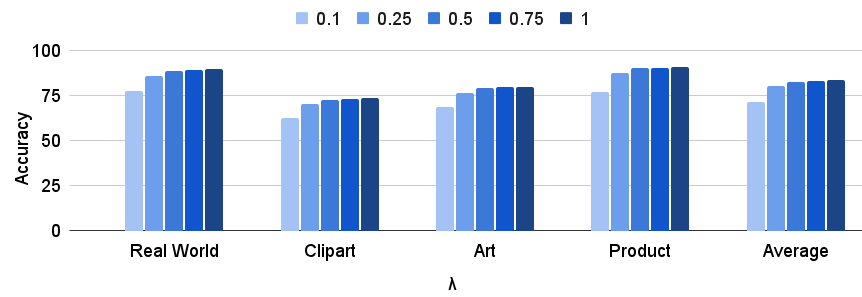}
    \caption{\textbf{Varying $\lambda$.} Performance of INDGIO on Office-Home (closed) when $\lambda$ in Equation 3 (main submission) is varied in $[0.1, 1.0]$.}
    \label{fig:hyperparam}
    \vspace{-6mm}
\end{figure}

\noindent \textbf{Training details.} The training is done on batch size of $240$ for 10 epochs and five runs. The visual branch is optimized using SGD optimizer with learning rate $1e-3$, weight decay $5e-5$, and nesterov momentum $0.9$. The fusion module uses same SGD optimizer but with learning rate $5e-3$. The learning rate is reduced by factor of $0.1$ after 6th epoch. In Equation 3 and 4 of main submission, the hyperparameter $\lambda$ is set to 1.0 for all experiments. In Figure \ref{fig:hyperparam} we vary $\lambda \in [0.1, 1.0]$ and observe that performance on average and across all  target domains increases with an increase in $\lambda$. This observation is consistent with our hypothesis as increasing $\lambda$ implies increasing the influence of intrinsic modality - which we leverage in this work.
%
%